\documentclass[11pt]{article}

\usepackage[preprint]{acl}

\usepackage{times}
\usepackage{latexsym}
\usepackage{xcolor}
\usepackage[T1]{fontenc}
\usepackage[utf8]{inputenc}

\usepackage{microtype}
\usepackage{inconsolata}

\usepackage{graphicx}
\usepackage{amsmath,amssymb}
\usepackage{todonotes}
\usepackage{booktabs}
\usepackage{multirow}
\usepackage{afterpage}
\usepackage{verbatim}

\renewcommand{\topfraction}{0.95}
\renewcommand{\dbltopfraction}{0.95}
\renewcommand{\bottomfraction}{0.05}
\renewcommand{\textfraction}{0.05}
\renewcommand{\floatpagefraction}{0.5}
\renewcommand{\dblfloatpagefraction}{0.5}

\newcommand{\safeincludegraphics}[2][]{%
  \IfFileExists{#2}{%
    \includegraphics[#1]{#2}%
  }{%
    \fbox{%
      \begin{minipage}[c][1.5in][c]{0.9\linewidth}
        \centering\small Missing figure: \texttt{\detokenize{#2}}
      \end{minipage}%
    }%
  }%
}

\title{Lost in Reconstruction: Aligning Action Representations with Language in Vision-Language-Action Models}

\author{
  Li Wenjie\thanks{~Correspondence to \texttt{wenjiel2@andrew.cmu.edu}.}
    \quad Yash Jangir \quad Ignacy Stepka \quad 
  \bfseries Yash Agarwal \quad Marion Kipsang \quad Yonatan Bisk \\
  Carnegie Mellon University \\}

\begin{document}
\maketitle

% Re-assert float-placement tweaks AFTER \maketitle since some classes reset
% counters at \begin{document}.
\setcounter{topnumber}{3}
\setcounter{dbltopnumber}{3}
\renewcommand{\topfraction}{0.95}
\renewcommand{\dbltopfraction}{0.95}
\renewcommand{\bottomfraction}{0.05}
\renewcommand{\textfraction}{0.05}
\renewcommand{\floatpagefraction}{0.4}
\renewcommand{\dblfloatpagefraction}{0.4}
\begin{abstract}
Action verbs describe not only the physical outcomes of actions, but also how those actions are performed. Yet action representations in vision-language-action models (VLAs) are typically optimized for reconstruction under L1/L2 losses in raw action space, where numerical proximity need not reflect linguistically meaningful distinctions. On BridgeV2, we show that action trajectories contain verb-grounding information beyond visual state changes, and that reconstruction-only discrete tokenization systematically erodes this information. To address this problem, we introduce \textbf{SALT}, a \textbf{S}emantically \textbf{AL}igned action \textbf{T}okenizer that augments a VQ-VAE-style tokenizer with an auxiliary objective requiring a frozen vision-language model to recover the episode instruction from quantized action latents. Policies trained with SALT achieve $71.9\%$ average success in SimplerEnv, compared with $42.7\%$ for a reconstruction-only VQ-VAE tokenizer and $31.2\%$ for FAST. SALT also develops verb-specialized codes while maintaining reconstruction fidelity. These results show that robot action trajectories provide a source of language grounding and that preserving this structure in action representations can substantially improve language-conditioned control.
\end{abstract}

% Verb-conditioned trajectories: shows both motion-dynamics and action-goal
% signals for five motion-shape verbs. Placed early (right after abstract) so
% it floats to the top of page 2 in two-column ACL formatting.
\begin{figure*}[!t]
  \centering
  \safeincludegraphics[width=\textwidth]{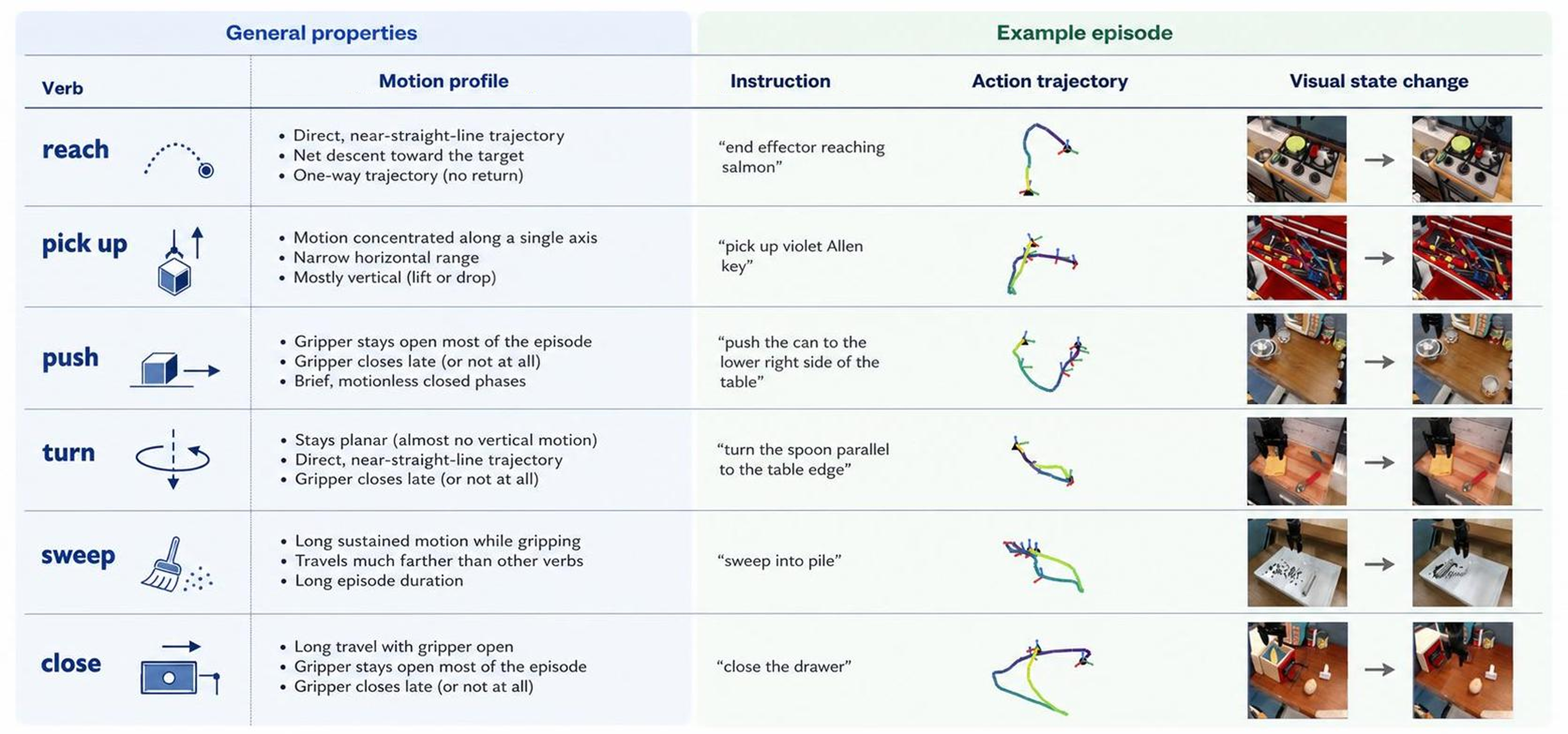}
  \caption{\textbf{Verbs describe both motion dynamics and action goal.}
  Six BridgeV2 verbs with their characteristic motion profile (trajectory
  shape, axis, gripper timing) and one example episode each (instruction,
  3D end-effector trajectory, first/last frames). Profiles are derived from action trajectories of the verb, using pre-defined formulas. Recipe in (Appendix~\ref{app:verb_features}).}
  \label{fig:verb_grid}
\end{figure*}

% SALT pipeline overview (hero figure). Declared back-to-back with the
% verb-grid figure so both queue early; with \dbltopfraction=0.95, LaTeX
% places them on consecutive page tops (p2 + p3).
\section{Introduction}

How is language grounded in embodied experience in the world? How much of it can be explained by visual input alone? In particular, do the action modality provides extra grounding signals for verbs? Verb meanings reflect regularities in how agents act in the world, such as patterns of motion, contact, and resulting changes in physical state. In turn, language shapes how actions are represented: which variations are grouped as instances of the same action class and which distinctions are treated as consequential. An action representation should not only be expressive enough to support execution, but also preserve linguistically meaningful distinctions.

However, action representations in existing vision-language-action models (VLAs)~\citep{brohan2023rt1,brohan2023rt2,kim2024openvla,black2026pi0} are not explicitly designed to be aligned with language abstractions. VLAs are trained with objectives and design choices defined primarily over the Euclidean physical control space, while their relationship to language is left for downstream policy training to infer. The implicit assumption is that having low action-space loss in predicting a trajectory also preserves the language grounding signals that connect the trajectory to the instruction that describes it.

This omission reflects how VLAs are constructed. VLAs adapt foundation vision-language models (VLMs)~\citep{vinyals2015show,karpathy2015deep,alayrac2022flamingo} for robot control by conditioning action prediction on visual observations and natural-language instructions. This paradigm allows robot policies to inherit representations already aligned across vision and language. In many existing VLAs and language-conditioned robotics literature, language serves primarily as goal conditioning. In particular, through attributes such as color, shape, or location, it usually identifies the object to be manipulated and specifies its desired location or physical state~\citep{shridhar2021cliport,jang2022bcz,mees2022calvin,brohan2023rt2,liu2023libero}. Because these aspects of an instruction can often be grounded in visual observations, alignment research in VLAs has concentrated largely on connecting referring expressions to objects and scenes~\citep{shridhar2021cliport,brohan2023rt2,kim2024openvla} and shaping visual representations through language supervision~\citep{nair2022r3m,karamcheti2023voltron,ma2023liv}. Action representations, which turn a VLM into a VLA, and their alignment with linguistic abstractions have received comparatively little attention.

We show that this mismatch creates a consequential bottleneck at the action interface of a VLA. Robot trajectories can be represented continuously \citep{black2026pi0}, or discretely as action tokens \citep{brohan2023rt1}. Discrete tokenization is especially common because it allows VLAs to reuse the autoregressive next-token prediction interface of pretrained VLMs. Moreover, it is sometimes used during pretraining before a model transitions to a continuous action head~\citep{black2026pi0}. Representative discrete action tokenization approaches include per-dimension \textbf{Bin} tokenization, used by RT-1, RT-2, and OpenVLA~\citep{brohan2023rt1,brohan2023rt2,kim2024openvla}; learned \textbf{VQ-VAE}-style tokenizers, optimized primarily for trajectory reconstruction~\citep{oord2018vqvae,wang2025vqvla,mete2024quest,liu2026oat}; and \textbf{FAST}, which transforms trajectories into frequency coefficients and compresses them through byte-pair encoding~\citep{pertsch2025fast}. In common practice, the tokenizer and the VLA are trained separately: tokenizer training determines how continuous actions are represented as discrete tokens, while VLA training determines, given vision and language, which token to predict. The action vocabulary is fixed before language enters the pipeline, and none of these approaches explicitly requires it to preserve linguistically meaningful distinctions.

Studying alignment between language and action representations requires realistic actions paired with naturalistic human language. We use BridgeV2~\citep{walke2023bridgedata}, a dataset of teleoperated real-robot manipulation trajectories annotated with free-form natural-language instructions. While bounded by the affordances of a WidowX robot arm and its 7-Degrees-of-Freedom (DoF) control space, BridgeV2 provides human-like action and language use, which most existing scripted pick-and-place robotics demonstrations do not provide. On this dataset, we establish three findings:

\textbf{(i) Action trajectories contain verb-grounding information that visual outcomes alone do not capture.} We decompose verb information into \textit{action goals} (the visual change between episode endpoints) and \textit{motion dynamics} (the 7-DoF action trajectory). Each carries verb information not captured by the other as verbs are grounded not only in resulting state changes, but also in motion profiles, such as contact patterns and gripper timing (Section~\ref{sec:motion_goal}). This demonstrates that grounding language for control requires alignment with both vision and action.

\textbf{(ii) Reconstruction-only action tokenization systematically erodes verb-grounding signals.} Across Bin, VQ-VAE, and FAST tokenizers, mutual information between verbs and tokenized action representations decreases, with larger losses under stronger compression (Section~\ref{sec:analysis}). Downstream language-conditioned policy training does not fully recover the missing structure. These results identify the discrete action interface as a bottleneck between language and control.

\textbf{(iii) Semantically aligned tokenization improves task performance and produces more linguistically meaningful action representations.} We introduce \textbf{SALT}, a \textbf{S}emantically \textbf{AL}igned action \textbf{T}okenizer that augments VQ-VAE training with an auxiliary instruction-generation objective (Section~\ref{sec:method}). During tokenizer training, quantized action latents are fed to a frozen pretrained VLM, which must generate the corresponding episode instruction. This supervision encourages actions described by similar language to receive similar representations while still preserving reconstruction fidelity in continuous control space.

In SimplerEnv~\citep{li2024simpler}, we empirically demonstrate that SALT substantially improves both semantic alignment and closed-loop task success (Section~\ref{sec:experiments}). Policies trained with SALT achieve a $71.9\%$ average task success rate, compared with $42.7\%$ for a reconstruction-only VQ-VAE tokenizer and $31.2\%$ for FAST. Although the alignment objective does not explicitly identify verbs, the learned vocabulary becomes organized around verb-relevant distinctions: individual codes become highly selective for action classes such as \emph{flip}, which reconstruction-only tokenizations distribute across generic, semantically mixed codes. 

Together, these results support a general design principle for language-conditioned control: \textbf{action representations should preserve not only executable trajectories but also the abstractions through which language represents action.}

\begin{figure*}[!t]
  \centering
  \safeincludegraphics[width=0.88\textwidth]{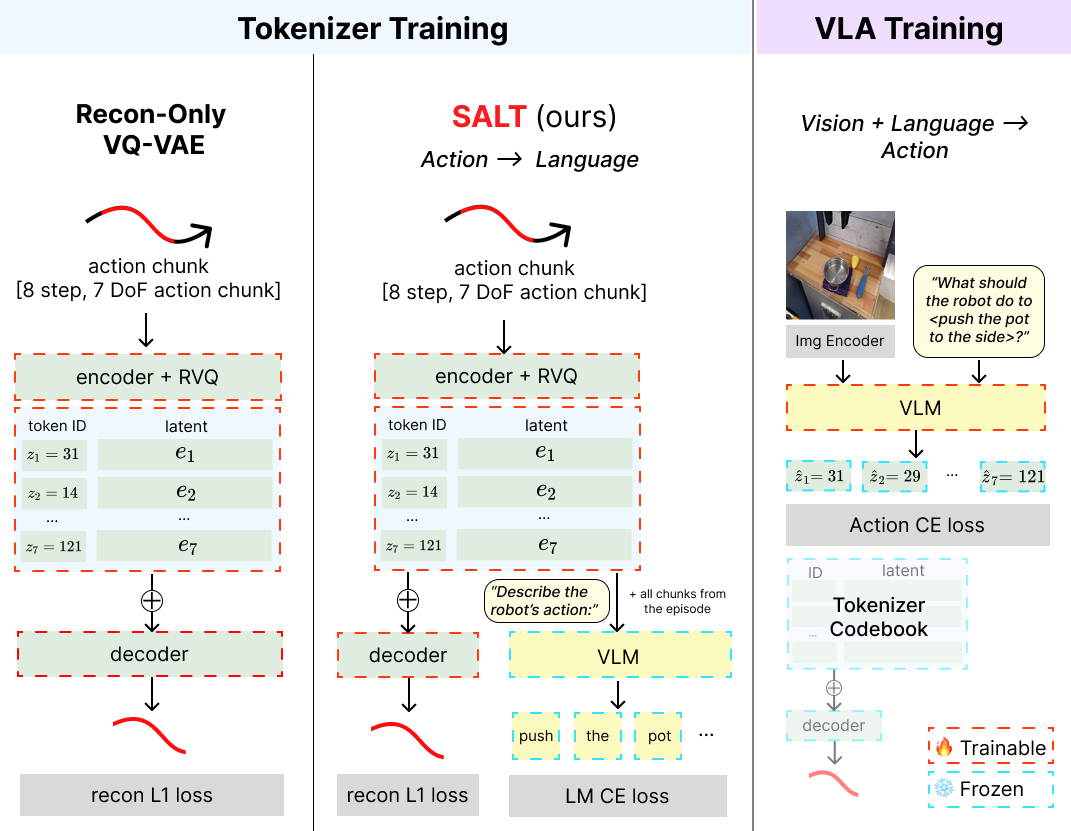}
  \caption{\textbf{SALT: semantically aligned action tokenization for
  VLAs.} \textbf{Left:} a reconstruction-only VQ-VAE tokenizer encodes
  an $8$-step, $7$-DoF action chunk with a residual-VQ encoder and is
  trained on reconstruction alone. \textbf{Middle (SALT):} the same
  architecture adds a generative alignment pressure: the quantized
  latents of all chunks in an episode are fed, with a describe prompt,
  to a frozen VLM that must generate the episode's instruction, and the
  LM cross-entropy shapes the encoder and codebook through the
  straight-through quantizer. \textbf{Right:} during VLA training, the VLM backbone predicts action-token IDs from vision and
  language (action cross-entropy), and the frozen tokenizer codebook
  and decoder turn predicted IDs into executable actions.}
  \label{fig:hero}
\end{figure*}

% ============================================================
\section{Diagnostic 1: Verbs Describe Both Action Goals and Motion Dynamics}
\label{sec2}
\label{sec:motion_goal}
% ============================================================

Language instructions can specify both an \emph{action goal}, which is the desired change in the physical environment, and the \emph{motion dynamics} through which that change is produced~\citep{hwang2024motif,zhang2026molmomotion}. This distinction parallels a long-standing observation in lexical semantics: verb meanings lexicalize \emph{manner} --- how an action is performed --- and/or \emph{result} --- the change it brings about~\citep{talmy1985lexicalization,levin1991wiping,rappaporthovav2010manner}. We operationalize the action goal as the visual change between an episode's first and last frames, and motion dynamics as its 7-DoF action trajectory. This distinction maps onto different VLA modalities, as the goal information can be conveyed through vision, whereas motion-specific information must enter through the action representation. Figure~\ref{fig:verb_grid} illustrates how verbs such as \emph{push}, \emph{flip}, and \emph{fold} are associated with characteristic patterns of translation, rotation, contact, and gripper timing.

We study this distinction on BridgeV2~\citep{walke2023bridgedata}, which pairs teleoperated real-robot manipulation trajectories with free-form natural-language instructions. After extracting and lemmatizing instruction verbs, we retain $17$ verb classes spanning $27{,}271$ episodes. Dataset processing and the motion-profile features shown in Figure~\ref{fig:verb_grid} are detailed in Appendices~\ref{app:dataset} and~\ref{app:verb_features}.

\paragraph{Estimating verb information.}
Let $Y$ denote the verb and $X$ denote motion dynamics, action goal, or both. We estimate
\begin{equation}
I(Y;X)=H(Y)-H(Y\mid X)
\end{equation}
from the held-out cross-entropy of matched Transformer classifiers under $5$-fold stratified cross-validation. The motion condition receives the 7-DoF action sequence, the goal condition receives frozen DINOv2-S~\citep{oquab2024dinov2} representations of the first and last frames, and the combined condition receives both. The architecture and optimization details are provided in Appendix~\ref{app:probe_details}, with a corroborating $R^2$ commonality analysis in Appendix~\ref{app:r2_commonality}.

\begin{table}[t]
\centering
\caption{\textbf{Action goals and motion dynamics provide complementary verb-grounding information on BridgeV2.}
Each entry reports the estimated contribution of a verb to
$I(Y;X)$ under the motion-only, goal-only, and combined conditions.
$\Delta_{\text{motion}}=\text{Both}-\text{Goal}$ denotes information
contributed uniquely by motion. A verb can carry unique information in both dimensions. The complete table appears in
Appendix~\ref{app:per_class}.}
\label{tab}
\label{tab:motion_goal_mi}
\small
\resizebox{\columnwidth}{!}{%
\begin{tabular}{lrrrrr}
\toprule
$Y=$ Verb & Motion & Goal & Both &
$\Delta_{\text{motion}}$ & $\Delta_{\text{goal}}$ \\
\midrule
\texttt{move} & 0.157 & 0.189 & 0.213 & $\mathbf{+0.023}$ & $\mathbf{+0.056}$ \\
\texttt{put}  & 0.134 & 0.143 & 0.161 & $\mathbf{+0.018}$ & $\mathbf{+0.027}$ \\
\texttt{fold} & 0.143 & 0.189 & 0.187 & $-0.001$ & $\mathbf{+0.044}$ \\
\texttt{push} & 0.011 & 0.008 & 0.014 & $+0.005$ & $+0.002$ \\
\bottomrule
\end{tabular}
}
\end{table}

\paragraph{Motion dynamics provide unique verb-grounding information.}
Table~\ref{tab:motion_goal_mi} shows that action goals and motion dynamics provide complementary information about verbs. In aggregate, the combined representation contains more verb information than either modality alone: motion contributes an estimated $0.059$ bits beyond the visual goal representation, while action goals contribute $0.282$ bits beyond motion. Visual state change explains more unique verb information overall, but does not exhaust the information available in the trajectory.

The motion-specific contribution is concentrated among verbs whose endpoint states can appear similar despite systematic differences in how the action is executed. For example, \texttt{move} and \texttt{put} together account for approximately two-thirds of the estimated motion-unique signal, while the less frequent verb \texttt{push} also contributes positively. Conversely, verbs whose meanings are strongly reflected in the resulting physical state are more strongly associated with the action-goal representation. For \texttt{fold}, for instance, the visual transition from an unfolded object to a folded configuration is particularly salient, while the estimated motion-unique contribution is negligible.

These results establish that verbs are grounded in both what an action achieves and how it is performed. Because the motion-specific component can reach a VLA only through its action representation, language-conditioned control requires action--language alignment in addition to vision--language alignment. A complete breakdown across all $17$ verbs is provided in Appendix~\ref{app:per_class}.

% Rate-distortion sweep (column-width). Placed at the start of §3 so it
% floats to the page of or just after the §3 title.
\begin{figure}[!t]
  \centering
  \safeincludegraphics[width=\columnwidth]{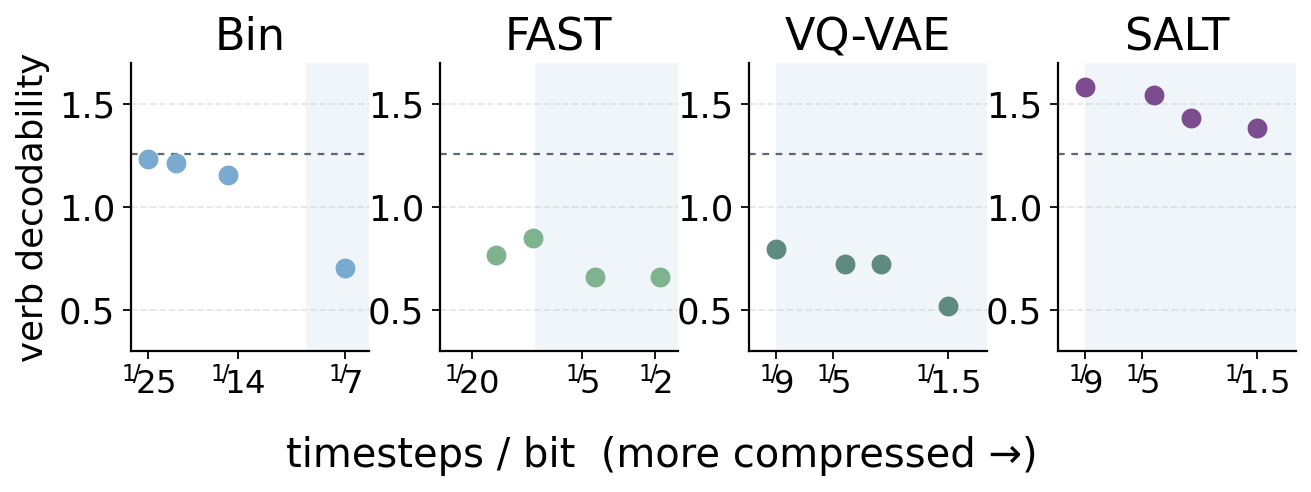}
  \caption{\textbf{A rate-distortion view of verb decodability erosion, across
  four tokenizer families on BridgeV2.} Verb decodability is the mutual information $I(\text{verb}; \text{tokens})$ in
  bits; the x-axis is timesteps per bit, so rightward is more
  compressed (a tick $1/N$ denotes $N$ bits per timestep;
  Appendix~\ref{app:bpts}). Dashed line: continuous-trajectory reference ($1.26$ bits). Every
  reconstruction-only tokenizer is below the reference and declines
  as compression grows. SALT closes the tokenization gap across
  the full compression range.
 }
  \label{fig:sweep_combined}
\end{figure}

% ============================================================
\section{Diagnostic 2: Semantic Loss in Discrete Tokenization}
\label{sec:analysis}
% ============================================================

Many VLAs represent continuous robot trajectories as discrete action tokens that can be predicted through the autoregressive interface of a pretrained VLM~\citep{brohan2023rt1,brohan2023rt2,pertsch2025fast,wang2025vqvla}. An action tokenizer $\tau$ maps each continuous action chunk $a_{1:H}\in\mathbb{R}^{H\times d_a}$ to a sequence of $K$ discrete symbols ($z_{1:K} \in \mathbb{V}^{K}$, where $\mathbb{V}$ is the token vocabulary):
$\tau: \mathbb{R}^{H \times d_a} \rightarrow \mathbb{V}^K$,
where $\tau(a_{1:H}) = z_{1:K}$.
We consider three representative families: \textbf{Bin}, which independently discretizes each action dimension; \textbf{FAST}~\citep{pertsch2025fast}, which transforms trajectories into frequency coefficients and compresses them with byte-pair encoding; and \textbf{VQ-VAE}~\citep{oord2018vqvae}, which learns an encoder, codebook, and decoder through trajectory reconstruction. Although these approaches partition action space differently, none explicitly optimizes the resulting representation to preserve distinctions expressed by language.

We apply the verb-decoding analysis from Section~\ref{sec:motion_goal} to token sequences produced by each tokenizer across compression levels. Figure~\ref{fig:sweep_combined} reports mutual information between verbs and tokenized actions as a function of bitrate, with the continuous trajectory as a reference. Across all three families, tokenized representations retain less verb information than the continuous actions, and the gap generally widens as compression increases. 

These results show that the reconstruction-only objective does not guarantee preservation of language grounding signals in compressed action representations. A small error in Euclidean space could be consequential for verbs, and larger variations could be incidental and semantically equivalent, bottlenecking downstream VLA performance. Additional metrics and methodology details are provided in Appendices~\ref{app:bpts} and~\ref{app:vanilla_probe_mf1}.

% ============================================================
\section{Method: \texorpdfstring{\textcolor{red}{S}emantically \textcolor{red}{AL}igned \textcolor{red}{T}okenization}{Semantically ALigned Tokenization}}
\label{sec:method}
% ============================================================
\textbf{SALT} augments a VQ-VAE action tokenizer \citep{wang2025vqvla} with language supervision, encouraging its latent space to preserve distinctions predictive of natural-language instructions. SALT intervenes exclusively in the tokenizer-training stage: it changes how the action vocabulary is learned, while VLA training and the policy's use of the tokens remain unchanged (Figure~\ref{fig:hero}). We focus on VQ-VAE tokenizers because, unlike fixed Bin or FAST representations, their encoder and codebooks can be directly shaped during training.

We instantiate SALT using a residual VQ-VAE. Given an action chunk $a^{(i)}_{1:H}$, the tokenizer predicts $K$ codebook indices $z_{i,1:K}$. The corresponding codebook vectors are summed to form the quantized latent

\begin{equation}
\mathbf{q}_i
=
\sum_{k=1}^{K}
\mathbf{e}^{(k)}_{z_{i,k}}.
\end{equation}

SALT augments the standard VQ-VAE objective with a language-alignment loss,

\begin{equation}
\mathcal{L}
=
\mathcal{L}_{\mathrm{recon}}
+
\mathcal{L}_{\mathrm{VQ}}
+
\lambda \mathcal{L}_{\mathrm{align}},
\end{equation}

where $\mathcal{L}_{\mathrm{recon}}$ is the action reconstruction loss, $\mathcal{L}_{\mathrm{VQ}}$ comprises the codebook and commitment losses \citep{oord2018vqvae}, and $\mathcal{L}_{\mathrm{align}}$ encourages the tokenizer to preserve information predictive of language.

For alignment, an episode is divided into $M$ action chunks, yielding quantized latents
$\mathbf{q}_{1:M} = (\mathbf{q}_1,\ldots,\mathbf{q}_M)$. We match the tokenizer latent dimension to the language-model embedding dimension and convert each latent into a soft prefix embedding,

\begin{equation}
\mathbf{p}_i
=
g\,\mathbf{q}_i
+
\mathrm{PE}(i),
\end{equation}

where $g$ is a learned scalar gain and $\mathrm{PE}(i)$ is a positional encoding. The resulting prefix sequence $\mathbf{P}=[\mathbf{p}_1,\ldots,\mathbf{p}_M]$
is provided, together with a short textual prompt $s$, to a frozen pretrained language model ($p_{\mathrm{LM}}$), which predicts the episode instruction $w_{1:L}$ where $L$ is the instruction length. The alignment objective is

\begin{equation}
\mathcal{L}_{\mathrm{align}}
= -\frac{1}{L} \sum_{t=1}^{L} \log p_{\mathrm{LM}}
\left( w_t \mid w_{<t},\mathbf{P},s\right).
\end{equation}

The language model remains frozen, while gradients propagate through the input embeddings and the straight-through quantizer into the tokenizer encoder and codebooks. Consequently, the alignment objective changes how the tokenizer partitions action space without modifying the downstream VLA architecture. Because it operates directly on free-form instructions, it requires neither a predefined verb inventory nor a separate text encoder or contrastive negative pairs.

After tokenizer training, the language model is discarded and the tokenizer is frozen. Downstream VLA training proceeds normally, that is the policy predicts discrete codebook indices, which are decoded by the tokenizer into continuous actions.

% ============================================================
\section{Experiments}
\label{sec:experiments}
% ============================================================

\subsection{Setup}
\label{sec:exp_setup}
% subsection: baseline, SimplerEnv
% result in a new section
We evaluate SALT as a drop-in action tokenizer for
miniVLA~\citep{belkhale2024minivla}, a Prismatic-style VLA with a
Qwen2.5-0.5B~\citep{qwen2025qwen25} backbone. All policies are trained
on BridgeV2 from the base vision-language checkpoint (no prior
robot-action pretraining) for $15$k gradient steps at a global batch
size of $128$. The action tokenizer is the only component that differs
across conditions.

We compare three tokenizers. \textbf{SALT} and \textbf{VQ-VAE} share
the same residual-VQ architecture ($8$-timestep chunks; $7$ residual
groups of $256$ codes each, so every chunk becomes $7$ token IDs) and
the same training data, and differ only in the language-alignment loss
of Section~\ref{sec:method} (VQ-VAE is trained with reconstruction
and commitment losses alone). \textbf{FAST}~\citep{pertsch2025fast} is
fitted on the same BridgeV2 chunks with vocabulary $1{,}024$ and
${\approx}7$ tokens per chunk. These hyperparameters are selected so
that all three tokenizers operate at a comparable compute and
compression budget: ${\approx}7$ tokens per $8$-timestep chunk, i.e.\
$7.0$--$8.6$ bits per timestep (Appendix~\ref{app:bpts}).

Rollout success is measured in SimplerEnv's visual-matching WidowX
suite~\citep{li2024simpler}: four tabletop tasks (put spoon on towel,
put carrot on plate, stack green block on yellow, put eggplant in
basket), $24$ episodes each ($96$ rollouts per policy), with open-loop
execution of $8$-step action chunks.

\begin{figure*}[t]
  \centering
  \safeincludegraphics[width=\textwidth]{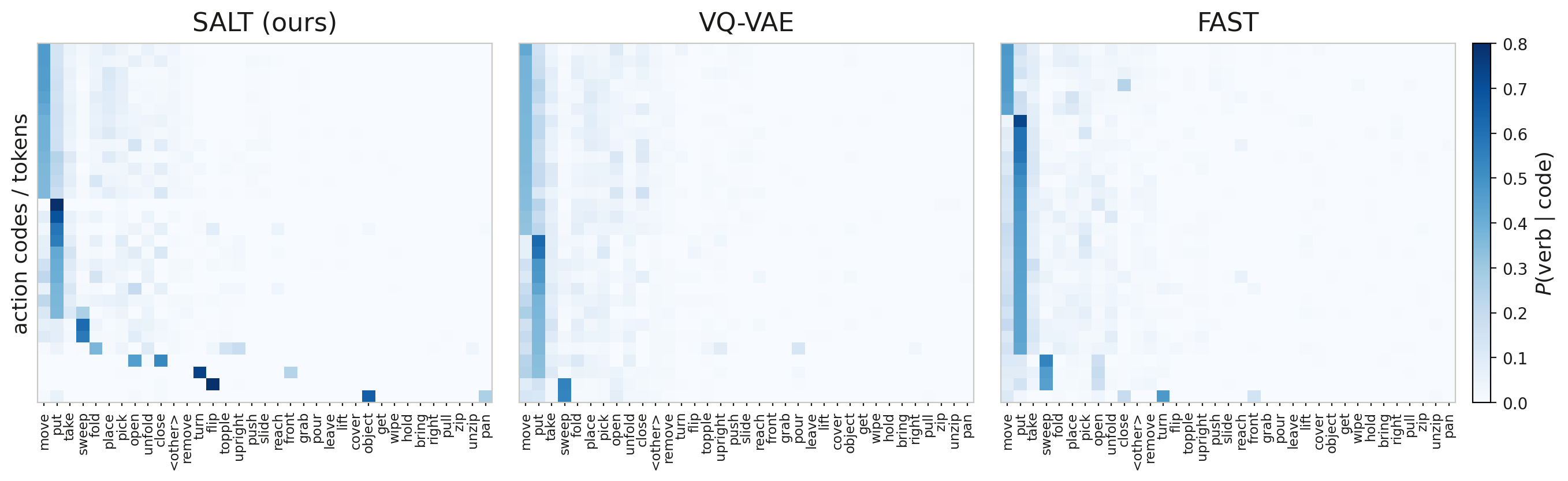}
  \caption{\textbf{Per-code verb distributions, first residual group.}
  For each tokenizer we show its $30$ most verb-selective units ---
  highest $\max_v P(\text{verb} \mid \text{code})$ among codes/tokens
  used in ${\ge}100$ training chunks --- as rows sorted by dominant
  verb; each cell is $P(\text{verb} \mid \text{code})$ by counting.
  Sharp verb-selective rows emerge for SALT across many verbs,
  including rare ones (\emph{flip} $98\%$, \emph{turn} $74\%$,
  \emph{pour}, \emph{topple}); the selective units of VQ-VAE (middle)
  and FAST (right) are confined to the most frequent verbs
  (\emph{put}, \emph{sweep}), with the remainder diffuse.}
  \label{fig:cooc_heatmap}
\end{figure*}

\begin{figure*}[t]
  \centering
  \safeincludegraphics[width=\textwidth]{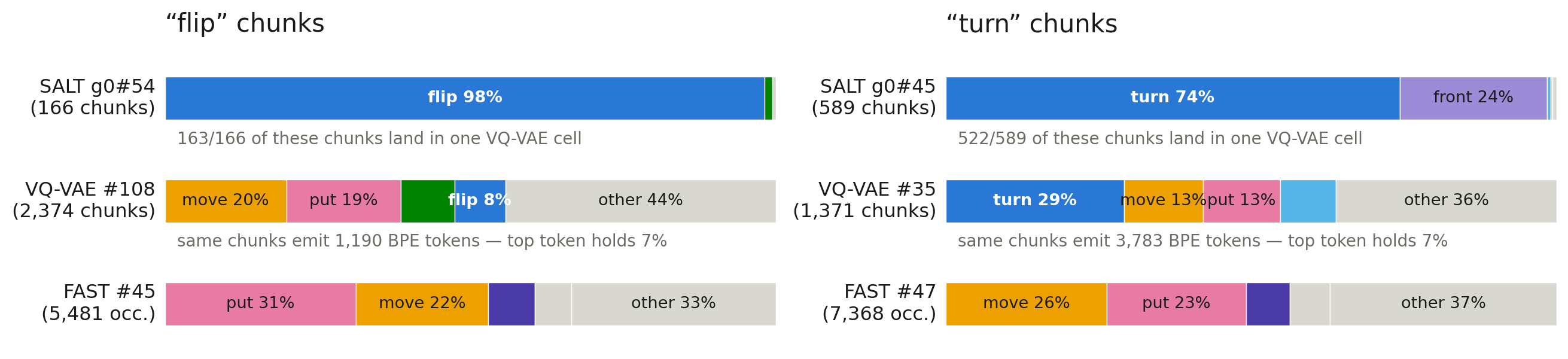}
  \caption{\textbf{The same action chunks under three tokenizations.}
  For SALT's \emph{flip}- and \emph{turn}-dedicated codes, the verb
  composition of (top bar) the SALT code, (middle) the single VQ-VAE
  cell that receives almost all of the same chunks, and (bottom) the
  FAST BPE token receiving the largest share of their token
  occurrences. SALT dedicates a symbol; VQ-VAE lumps the chunks into
  a generic mixed cell; FAST distributes them across generic tokens
  with no dominant home. The \emph{front} share of SALT's turn code
  comes entirely from the instruction \emph{``lever vertical to
  front''} --- the same lever-turning action worded without ``turn''
  --- so the code unites this paraphrase with \emph{turn}: the
  alignment tracks meaning, not surface wording.}
  \label{fig:cooc_case}
\end{figure*}

\subsection{Results}
\label{sec:main_results}

\begin{table}[t]
  \centering
  \caption{SimplerEnv WidowX rollout success rate (\%) for
  BridgeV2-trained miniVLA, per task ($24$ episodes each) and averaged
  ($n{=}96$). Tasks: put spoon on towel, put carrot on plate, stack
  green block on yellow, put eggplant in basket. All policies are
  trained identically for $15$k steps; the action tokenizer is the only
  changed component. Best in \textbf{bold}.}
  \label{tab:bridge_main}
  \small
  \resizebox{\columnwidth}{!}{%
  \begin{tabular}{lccccc}
    \toprule
    Tokenizer & Spoon & Carrot & Stack & Eggplant & Mean $\uparrow$ \\
    \midrule
    FAST      & $54.2$ & $29.2$ & $20.8$ & $20.8$ & $31.2$ \\
    VQ-VAE  & $58.3$ & $45.8$ & $33.3$ & $33.3$ & $42.7$ \\
    SALT      & $\mathbf{75.0}$ & $\mathbf{62.5}$ & $\mathbf{70.8}$ & $\mathbf{79.2}$ & $\mathbf{71.9}$ \\
    \bottomrule
  \end{tabular}
  }
\end{table}

\begin{table}[t]
  \centering
  \caption{Verb decodability and reconstruction fidelity of the policy
  tokenizers (probe and protocol of Section~\ref{sec:analysis},
  $5$-fold stratified CV; accuracy companions in
  Appendix~\ref{app:vanilla_probe_mf1}). \emph{TokID} probes the
  discrete token IDs; $E_\text{in}$ probes the trained policy's frozen
  action-token input embeddings (fold $0$). \emph{Recon} is held-out
  reconstruction L1 in normalized action units. Best in
  \textbf{bold}.}
  \label{tab:vanilla_probe_mf1_main}
  \label{tab:transfer_probe}
  \small
  \resizebox{\columnwidth}{!}{%
  \begin{tabular}{lccc}
    \toprule
    Tokenizer & Recon $\downarrow$ & TokID MF1 $\uparrow$ & $E_\text{in}$ MF1 $\uparrow$ \\
    \midrule
    FAST ($V{=}1024$)            & $0.113$ & $30.3$ & $36.3$ \\
    VQ-VAE                     & $\mathbf{0.080}$ & $37.3$ & $38.3$ \\
    SALT                         & $0.088$ & $\mathbf{39.1}$ & $\mathbf{43.7}$ \\
    \midrule
    Native (continuous, ref.)    & ---     & $53.0$ & --- \\
    \bottomrule
  \end{tabular}
  }
\end{table}

\paragraph{SALT improves deployment success.}
Table~\ref{tab:bridge_main} reports rollout success. SALT reaches
$71.9\%$, against $42.7\%$ for VQ-VAE and $31.2\%$ for FAST, and
leads on every individual task, with the largest margins on the two
most difficult (stack: $70.8$ vs.\ $33.3$; eggplant: $79.2$ vs.\
$33.3$). The
SALT-vs-VQ-VAE comparison isolates the alignment loss: the two
tokenizers share architecture, capacity, data, and the entire VLA
training recipe, so the $29.2$-point gap is attributable to how the
action vocabulary is partitioned, not to how much it compresses.

\paragraph{Language alignment transfers beyond the tokenizer.}
Although the alignment objective supervises only the tokenizer's latent representations, its effects propagate to both the discrete token IDs produced by the tokenizer and the action-token embeddings learned by the downstream VLA (Table~\ref{tab:transfer_probe}). This transfer is not guaranteed: the alignment loss is applied to the quantized latents inside the tokenizer, and it never directly supervises (i) the discrete symbols or (ii) the action-token embeddings re-initialized during VLA training. Nevertheless, SALT achieves the highest verb decodability at both stages, reaching $39.1$ macro-F1 on token IDs (vs.\ $37.3$ for VQ-VAE and $30.3$ for FAST) and $43.7$ on learned action embeddings (vs.\ $38.3$ and $36.3$). Moreover, its token-ID accuracy ($58.7\%$) matches the continuous-action reference ($58.0\%$; Appendix~\ref{app:vanilla_probe_mf1}). These results suggest that semantic structure introduced during tokenizer training survives discretization and is preserved throughout downstream policy learning.

\paragraph{Semantic alignment preserves reconstruction fidelity.}
The improved semantic structure does not come at the expense of reconstruction quality. At the matched compression rate of seven tokens per eight-step chunk, SALT's held-out reconstruction error remains close to that of VQ-VAE (Table~\ref{tab:transfer_probe}). Likewise, the characteristic motion signatures identified in Section~\ref{sec:motion_goal} are largely preserved after reconstruction: the rank correlation of one-vs-rest effect sizes across $63$ interpretable trajectory features remains at least $0.92$ (vs.\ $0.96$ for VQ-VAE), and distinctive patterns such as \emph{flip}'s dominant rotational motion are retained. Thus, semantic alignment primarily changes how action trajectories are partitioned into discrete symbols rather than how faithfully those symbols reconstruct continuous actions.

\paragraph{Verb-specialized codes emerge under semantic alignment.}
Probe accuracy shows that SALT preserves more verb information, but does not reveal how its vocabulary is organized. We therefore examine code--verb co-occurrence directly, without training an additional classifier. If codes specialize by action semantics, each code's distribution should concentrate on one or a small number of verbs, producing sharp verb-selective bands; semantically mixed codes instead produce diffuse distributions. Figure~\ref{fig:cooc_heatmap} shows that SALT develops highly selective codes associated with particular verbs, whereas VQ-VAE distributes the same trajectories across semantically mixed codes and FAST largely reflects the corpus verb distribution. Figure~\ref{fig:cooc_case} illustrates this effect for representative \emph{flip} and \emph{turn} codes. Interestingly, the \emph{turn} code also captures instructions phrased as ``lever vertical to front,'' grouping semantically equivalent actions despite different wording. Beyond these examples, the same pattern holds quantitatively. A probe-free majority-vote lookup based solely on code--verb co-occurrence achieves higher held-out accuracy for SALT than VQ-VAE for every individual residual group and every cumulative group prefix (Appendix~\ref{app:cooc_lookup}); for example, using only the first two residual groups yields $46.3\%$ episode-level accuracy for SALT versus $43.6\%$ for VQ-VAE (McNemar $p=.011$), with FAST reaching $35.0\%$. Together, these results indicate that semantic alignment reorganizes the action vocabulary around linguistically meaningful action categories rather than reconstruction alone.

% ============================================================
\section{Related Work}
\label{sec:related}
% ============================================================
\paragraph{Embodied language grounding and action semantics.}
Embodied language grounding studies how words connect to perception,
affordances, skills, and action. Most robotics work grounds language in
objects, spatial relations, task goals, or feasible
plans~\citep{shridhar2021cliport, mees2022calvin, ahn2022icanisay}. We
focus on verbs because they provide a compact probe of action-level
structure: they group trajectories by motion pattern, contact mode,
gripper timing, force, path/result structure, and object interaction.
Unlike work aimed at learning better language representations, we use
language as supervision and diagnosis for learning better action
representations. Our representational diagnostics adapt the probing
methodology developed for
NLP~\citep{hewitt2019designing,belinkov2022probing} to the
discrete-token interface of a VLA.

\paragraph{Cross-modal alignment in robot learning.}
Most language-conditioned robot learning aligns language with perception. Methods such as CLIPort~\citep{shridhar2021cliport}, BC-Z~\citep{jang2022bcz}, CALVIN~\citep{mees2022calvin}, and SayCan~\citep{ahn2022icanisay} use language primarily to specify objects, spatial relations, or task goals. Our work instead studies alignment at the action interface, asking whether action representations themselves preserve the distinctions expressed by language.

\paragraph{Vision-language-action models.}
Vision-language-action (VLA) models adapt pretrained vision-language or language-model backbones for robot control by introducing an action prediction interface~\citep{brohan2023rt1,brohan2023rt2,driess2023palme,kim2024openvla}. While this paradigm benefits from pretrained visual and linguistic representations, it raises a representational question that has received comparatively little attention: how continuous robot actions should be represented before they are consumed by the backbone. Our work focuses specifically on this action interface rather than scaling VLA architectures themselves.

\paragraph{Action representations in VLAs.}
Existing VLAs represent actions either as discrete tokens or continuous action chunks. Discrete approaches, including RT-1/RT-2-style binning~\citep{brohan2023rt1,brohan2023rt2}, VQ-based tokenizers~\citep{lee2024vqbet,wang2025vqvla}, FAST~\citep{pertsch2025fast}, BeT~\citep{shafiullah2022bet}, QueST~\citep{mete2024quest}, LAPA~\citep{ye2024lapa}, and OAT~\citep{liu2026oat}, are typically optimized for reconstruction or self-supervised prediction. Continuous approaches such as Diffusion Policy~\citep{chi2024diffusionpolicy}, Octo~\citep{octomodelteam2024octo}, and $\pi_0$~\citep{black2026pi0} avoid discretization by predicting continuous action chunks. Our work addresses the discrete setting, showing that action tokenization should preserve language-relevant structure in addition to executable trajectories.

\paragraph{Language-aligned representations for robot control.}
A separate line of work uses natural language as supervision for
\emph{visual} representations in robotics. R3M~\citep{nair2022r3m}
aligns video frames with time-shifted language captions,
Voltron~\citep{karamcheti2023voltron} jointly reconstructs and grounds
language during manipulation pretraining, and LIV~\citep{ma2023liv}
unifies value learning with vision-language alignment via a
CLIP-style~\citep{radford2021clip} contrastive objective. These methods
inject language structure into the perception side of a robot policy.
SALT applies the analogous idea on the \emph{action} side: instead of
aligning a visual encoder with instructions, we align the latent space
of an action tokenizer with instructions, so that the discrete action
vocabulary itself carries language-relevant geometry. It also differs
in mechanism: where these methods use contrastive objectives against a
text encoder, SALT's pressure is generative because a frozen LM must
produce the instruction from the quantized action latents, which
aligns the representation to the same soft-token interface a VLA
backbone consumes.

\section*{Limitations}

Our conclusions are subject to several limitations. First, current language-conditioned robotics datasets exhibit limited linguistic diversity: after filtering, BridgeV2 contains only $17$ verb classes. Although sufficient to study action-language alignment, richer datasets with broader verb inventories would better test whether the benefits of semantic tokenization grow with linguistic diversity.

Second, SALT currently applies only to tokenizers with a learnable latent representation. Extending similar semantic supervision to fixed discretization schemes such as Bin or signal-processing approaches such as FAST remains an open question.

Third, our experiments use a relatively small VLA ($0.5$B parameters), a single training dataset, and simulation-based evaluation in SimplerEnv. Future work should determine whether the observed gains persist under large-scale multi-embodiment pretraining and real-robot deployment.

Finally, while we show that semantic alignment produces more interpretable action vocabularies and improves downstream policy performance, we do not establish a causal mechanism relating these two effects. Understanding how semantically organized action codes influence policy learning remains an important direction for future work.

% ============================================================
\bibliography{references}
% ============================================================
% \newpage
\appendix

\section{Dataset: BridgeV2}
\label{app:dataset}

We conduct all experiments on BridgeV2~\citep{walke2023bridgedata}, a large-scale
dataset of real-robot tabletop manipulation collected across multiple laboratory
environments.

\paragraph{Setting.}
Episodes are collected via human teleoperation of a WidowX~250 6-DoF robot arm
equipped with a parallel-jaw gripper.
The workspace is a tabletop with diverse household objects (toys, kitchenware,
cloths, blocks, etc.).
Each episode captures a single short-horizon manipulation task---typically
lasting ${\sim}37$ timesteps---from a third-person camera mounted above the table.

\paragraph{Observations and actions.}
At each timestep, the dataset records an RGB image
(\texttt{image\_0}, $256{\times}256$) and a 7-dimensional action vector:
\begin{equation}
  a_t = [\Delta x, \Delta y, \Delta z, \Delta \phi, \Delta \theta, \Delta \psi, g]
        \in \mathbb{R}^7,
\end{equation}
where $\Delta x{:}\Delta \psi$ are end-effector velocity commands
(3 translational + 3 rotational) and $g \in \{0, 1\}$ is the gripper open/close
command.
Each episode is thus a variable-length sequence of (image, action) pairs.

\paragraph{Instructions.}
Every episode is paired with a natural-language task instruction
(e.g., ``put the corn into the pot,'' ``fold the cloth in half'').
Instructions were provided by the teleoperators at collection time, describing the
intended task for each demonstration.
BridgeV2 contains 27{,}575 episodes (after basic quality filtering) spanning
519 unique manipulated objects.

\paragraph{Verb labels.}
We extract a primary verb from each instruction using SpaCy dependency parsing.
Verb forms are lemmatized (``moved'' $\to$ ``move''), directional variants are
merged (``put down/on/up/in'' $\to$ ``put''), and non-English or non-verb tokens
are dropped.
Applying a minimum-count filter ($\geq$30 training examples) yields
17 verb classes covering 27{,}271 episodes.
For VLA training we split 90/10 by episode into 24{,}544 training and 2{,}727
validation samples (seed = 42); for the probes of Sections~\ref{sec:motion_goal} and~\ref{sec:analysis} we use 5-fold stratified CV
with the same seed.

\begin{figure}[t]
  \centering
  \safeincludegraphics[width=\columnwidth]{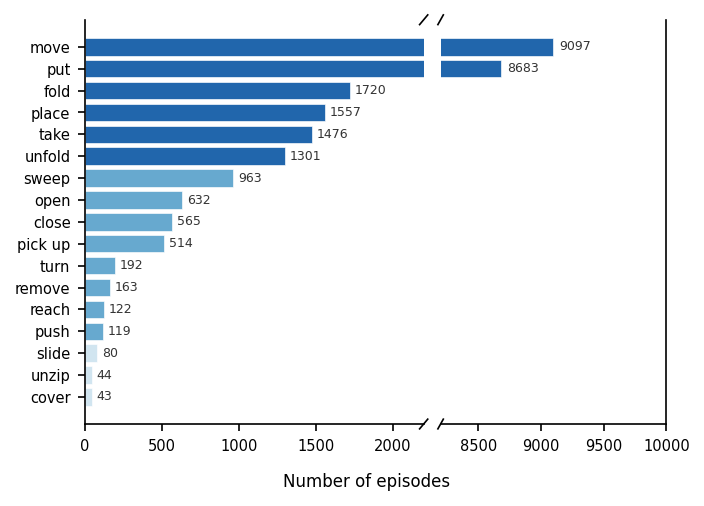}
  \caption{Distribution of the 17 verb classes in BridgeV2
    (episodes with $\geq$30 training examples).
    The two dominant verbs (\emph{move}, \emph{put}) comprise 65\% of
    episodes; the long tail spans 15 additional verbs.}
  \label{fig:verb_dist}
\end{figure}

Figure~\ref{fig:verb_dist} shows the resulting distribution.
The class frequencies are heavily imbalanced: the top two verbs (\emph{move},
\emph{put}) account for 65\% of all episodes, while the five rarest classes
each have fewer than 200 examples.
All probes are nonetheless trained with vanilla cross-entropy on the natural
class distribution (Appendix~\ref{app:probe_details}), which is required for
the calibrated posteriors the MI estimates rely on; class imbalance is
reported through macro-averaged metrics instead.

\section{Verb Motion-Profile Features}
\label{app:verb_features}

This appendix details how the per-verb ``characteristic motion
profiles'' of Figure~\ref{fig:verb_grid} are derived. The procedure is
fully automatic given the verb labels; no profile is hand-written.

\paragraph{Feature bank.}
Each episode's $7$-DoF action trajectory (no images) is summarized by
${\sim}63$ scalar features chosen to be interpretable and
scene-invariant --- they describe \emph{how} the arm moved, not
\emph{where} in the workspace. The features fall into eight families:
\textbf{(i) path geometry}: path length, net displacement, path
efficiency (net$/$total), mean curvature, direction changes, per-axis
totals $|\Delta x|, |\Delta y|, |\Delta z|$, dominant-axis fraction,
lateral-to-vertical ratio;
\textbf{(ii) rotation}: per-axis $|\Delta\text{roll/pitch/yaw}|$ and
signed totals, rotation-to-translation ratio, rotation dominant-axis
fraction;
\textbf{(iii) velocity-profile shape}: mean and peak speed, time of
peak speed, early-vs-late energy ratio, monotonicity, number of speed
peaks, kurtosis;
\textbf{(iv) acceleration and smoothness}: peak acceleration,
acceleration variance, mean jerk, mean cosine between consecutive
displacement vectors;
\textbf{(v) gripper state and timing}: open fraction,
close$\to$open transitions, time to first close/open, closed- and
open-phase durations, gripper state at peak speed;
\textbf{(vi) contact-conditioned motion}: path length and mean speed
while the gripper is closed vs.\ open, closed-to-open speed ratio,
vertical motion while closed;
\textbf{(vii) spatial extent}: per-axis bounding-box ranges,
bounding-box volume, radius of gyration;
\textbf{(viii) symmetry and oscillation}: start--end distance,
out-and-back overlap, dominant speed frequency, per-axis zero
crossings.

\paragraph{Scoring and selection.}
For each verb we sample up to $100$ episodes (seed $42$, minimum $5$
timesteps) from the same $17$-class corpus as
Section~\ref{sec:motion_goal}, compute all features, and score every
(verb, feature) pair with a one-vs-rest Cohen's $d$:
\begin{equation}
  d_{v,f} \;=\;
  \frac{\bar{x}_{v,f} - \bar{x}_{\lnot v,f}}{s_{\text{pooled}}},
\end{equation}
where $\bar{x}_{v,f}$ is the focal verb's mean on feature $f$,
$\bar{x}_{\lnot v,f}$ the pooled mean of all other verbs, and
$s_\text{pooled}$ the pooled standard deviation. A verb's
characteristic profile is its top-$3$ features by $|d|$, subject to
$|d| \ge 0.6$ (a medium-or-larger effect); each selected feature is
rendered in Figure~\ref{fig:verb_grid} with a plain-English label and
the sign of its deviation. The six verbs shown in the figure were
chosen to span distinct motion archetypes (approach, lift, push,
planar rotation, sustained wiping while gripping, fixture closing);
the same procedure applies to all $17$ classes.

\section{Bits per Timestep}
\label{app:bpts}

We use \emph{bits per timestep} (bpts) as the underlying rate measure
for the tokenizer sweeps. For a tokenizer that emits $K$ discrete
tokens drawn from a vocabulary of size $V$ per chunk of $T$ continuous
timesteps, the per-timestep rate is
\begin{equation}
  \text{bpts} \;=\; \frac{K \, \log_2 V}{T}.
\end{equation}
Figure~\ref{fig:sweep_combined} plots its inverse, \emph{timesteps per
bit} ($1/\text{bpts}$), so that the axis grows with compression: an
episode segment of $N$ timesteps that tokenizes into $N$ bits sits at
$1$, and more aggressive tokenizers sit further right. Axis ticks are
labeled $1/N$, i.e.\ $N$ bits per timestep.
This is the average number of bits the tokenized stream carries per
original action timestep. For variable-length tokenizers (e.g.\ FAST)
$K$ is replaced by its empirical mean over the training set. Bpts is
meaningful for comparing configurations \emph{within} a tokenizer
family (Bin sweep with varying bin counts, FAST sweep with varying
vocabulary or scale parameter, VQ-VAE sweep with varying codebook
size); cross-family absolute positions reflect architectural rate
ranges (e.g.\ Bin emits 7 tokens per timestep regardless of bin count;
VQ-VAE emits 1 token per chunk) and should not be over-interpreted as
a global compression ordering.

\section{Macro-F1 Version of Tokenizer Probes}
\label{app:vanilla_probe_mf1}

Table~\ref{tab:vanilla_probe_mf1} tabulates the Section~\ref{sec:analysis}
token-ID verb probes at representative configurations, with overall accuracy
(Acc) and per-class accuracy (PCA) reported alongside macro-F1.

\begin{table}[htbp]
\centering
\caption{Token-ID verb probes on BridgeV2 (\%; $5$-fold stratified CV;
standard errors $\le 1.5$ MF1). Accompanies
Section~\ref{sec:analysis} and Figure~\ref{fig:sweep_combined}.}
\label{tab:vanilla_probe_mf1}
\resizebox{\columnwidth}{!}{%
\begin{tabular}{lccc}
\toprule
Tokenizer                  & Acc  & PCA  & MF1  \\
\midrule
Bin (256/dim)              & $50.7$ & $42.0$ & $43.1$ \\
FAST (vocab $= 256$)       & $51.1$ & $33.6$ & $34.8$ \\
VQ-VAE (ng9/nemb256)       & $52.7$ & $30.3$ & $34.8$ \\
\midrule
Native (continuous, ref.)  & $58.0$ & $52.3$ & $53.0$ \\
\bottomrule
\end{tabular}
}
\end{table}

Table~\ref{tab:transfer_probe_acc} is the same companion for the
token-ID probes of the three Section~\ref{sec:experiments} tokenizers
(identical protocol; the VQ tokenizers emit $7$ residual-group IDs per
$8$-step chunk, and FAST is the policy's fitted vocabulary-$1{,}024$
tokenizer applied per chunk).

\begin{table}[htbp]
\centering
\caption{Accuracy companion for the Section~\ref{sec:experiments}
token-ID probes (\%; $5$-fold stratified CV; per-fold sd $\le 1.5$).
Accompanies Table~\ref{tab:transfer_probe}.}
\label{tab:transfer_probe_acc}
\resizebox{\columnwidth}{!}{%
\begin{tabular}{lccc}
\toprule
Tokenizer                  & Acc  & PCA  & MF1  \\
\midrule
FAST (chunked, vocab $=1024$) & $50.7$ & $28.8$ & $30.3$ \\
VQ-VAE                   & $54.5$ & $36.7$ & $37.3$ \\
SALT                       & $58.7$ & $37.9$ & $39.1$ \\
\midrule
FAST $E_\text{in}$ (fold $0$)     & $55.7$ & $35.4$ & $36.3$ \\
VQ-VAE $E_\text{in}$ (fold $0$) & $54.1$ & $39.0$ & $38.3$ \\
SALT $E_\text{in}$ (fold $0$)     & $62.4$ & $43.3$ & $43.7$ \\
\midrule
Native (continuous, ref.)  & $58.0$ & $52.3$ & $53.0$ \\
\bottomrule
\end{tabular}
}
\end{table}

\section{Held-Out Code--Verb Lookup Accuracy}
\label{app:cooc_lookup}

\begin{figure*}[t]
  \centering
  \safeincludegraphics[width=0.92\textwidth]{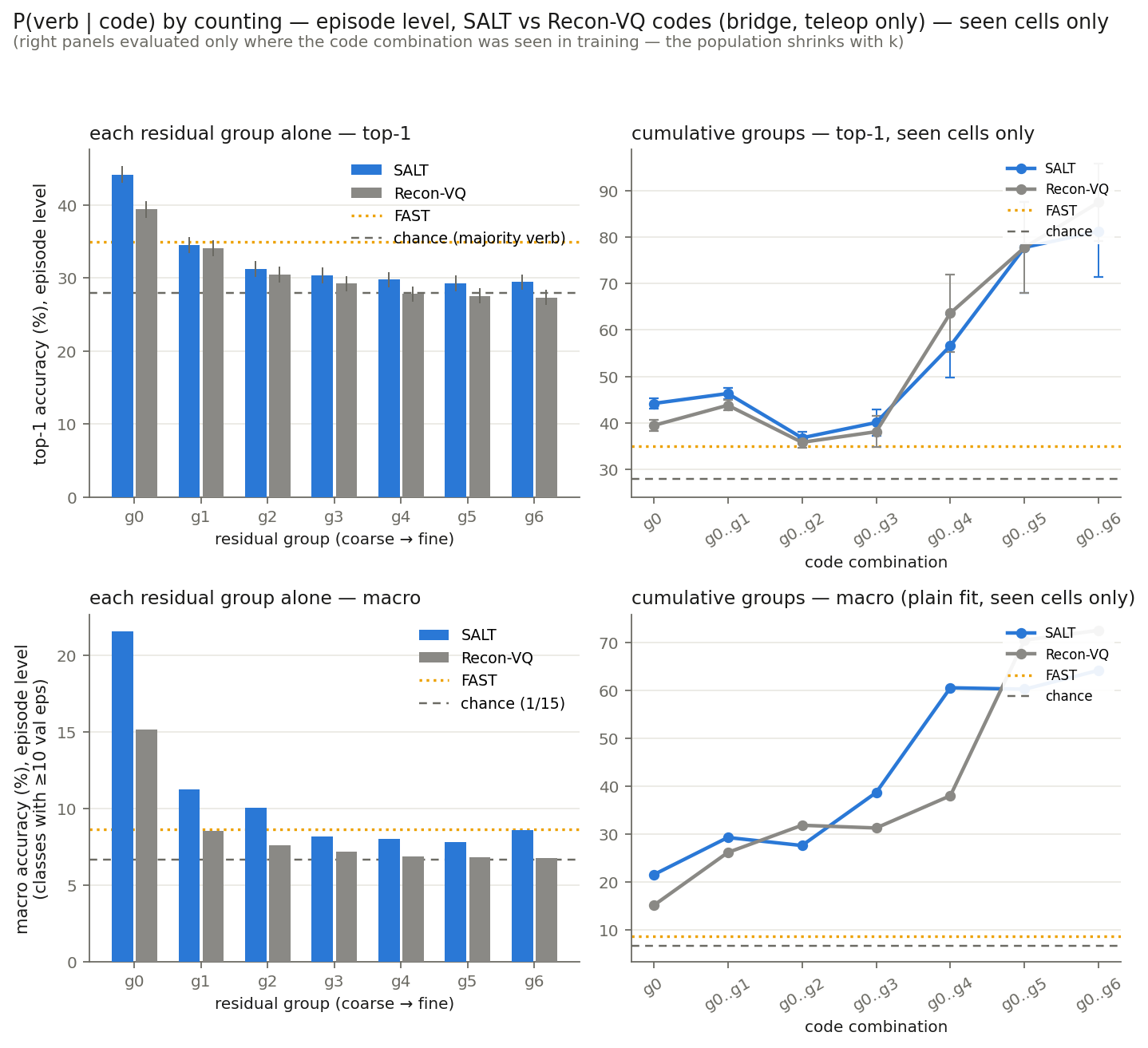}
  \caption{\textbf{$P(\text{verb}\mid\text{code})$ by counting:
  majority-vote lookup accuracy on held-out episodes (seen code
  combinations).} Top row: top-1 accuracy; bottom row: macro accuracy
  over classes with ${\ge}10$ validation episodes. Left: each residual
  group alone; right: cumulative group prefixes. SALT exceeds VQ-VAE
  at every group and prefix; FAST references dotted, chance dashed.}
  \label{fig:cooc_lookup}
\end{figure*}

Figure~\ref{fig:cooc_lookup} details the probe-free lookup statistic
referenced in Section~\ref{sec:main_results}. On training chunks we
record, for each code (or combination of codes), the majority verb of
the episodes it fires in; on held-out episodes we predict by soft
voting over each episode's chunks and score accuracy. The evaluation
is restricted to code combinations seen in training. The top row
reports top-1 accuracy; the bottom row macro-averages over verb
classes with at least $10$ validation episodes. SALT exceeds VQ-VAE
for every individual residual group and every cumulative group prefix.

\section{Probe Architecture and Training Details}
\label{app:probe_details}

This appendix details the verb classifier used in Section~\ref{sec:motion_goal} to estimate $H(Y \mid X^{(m)})$ for $m \in \{a, v, (a,v)\}$.

\paragraph{Architecture.}
A single Transformer encoder consumes a sequence
\[
    [\mathrm{CLS},\; v_1,\ldots,v_{V},\; a_1,\ldots,a_{T}],
\]
where $v_i$ are vision patch tokens (frozen DINOv2-S patch features projected to $d_{\rm model}$, two endpoint frames $\Rightarrow V = 2 \cdot 49 = 98$ tokens) and $a_t$ are action tokens (linear projection of the 7-D action vector at timestep $t$, padded to a fixed batch length). Sinusoidal temporal positional encoding on action tokens, learned 2-D positional encoding on patch tokens, and learned modality-type embeddings added to vision and action token streams. The encoder is pre-norm with $4$ layers, $d_{\rm model} = 128$, $8$ attention heads, dim-feedforward $4 d_{\rm model}$, and dropout $0.1$. The classifier head reads the $\mathrm{CLS}$ token and applies $\mathrm{LayerNorm} \to \mathrm{ReLU} \to \mathrm{Dropout} \to \mathrm{Linear}(d_{\rm model}, |\mathcal{Y}|)$, with $|\mathcal{Y}| = 17$.

\paragraph{Three input variants.}
The same architecture supports three masking conditions selected at forward time: motion-only masks the $V$ vision tokens out of attention, goal-only masks the $T$ action tokens, and both leaves the full sequence unmasked. The CLS token always attends to whatever is unmasked. Since masked tokens contribute no gradients, this is equivalent to training three separate probes whose only architectural difference is which tokens they receive --- the parameter count is identical across the three runs.

\paragraph{Training.}
Vanilla cross-entropy on the natural verb distribution (no class weighting, no label smoothing); this is the right loss for a calibrated $p(y \mid x)$ estimate, which is what the variational MI bound requires. AdamW with learning rate $10^{-4}$, batch size $32$, weight decay $10^{-2}$, OneCycleLR scheduler, gradient clipping at $1.0$. Up to $100$ epochs with early stopping on val cross-entropy, patience $15$. The model checkpoint with the lowest val cross-entropy is used for held-out logit extraction.

\paragraph{Cross-validation.}
Stratified $5$-fold CV on the $27{,}271$ verb-labeled BridgeV2 episodes (\texttt{StratifiedKFold}, seed $42$). The same fold assignment is used across the three modality conditions, so the per-fold differences $H_k(Y \mid X_v) - H_k(Y \mid X_a, X_v)$ are paired samples. Across the $5$ folds, every episode appears in val exactly once.

\paragraph{Estimating MI.}
For each $(m, k)$ we extract logits on fold $k$'s held-out episodes, apply log-softmax, and compute
\[
    \hat H_k(Y \mid X^{(m)}) \;=\; -\frac{1}{N_k} \sum_{i \in \mathrm{val}_k} \log_2 p_\theta(y_i \mid x_i^{(m)}),
\]
in bits. Then $\hat I_k(Y; X^{(m)}) = \hat H(Y) - \hat H_k(Y \mid X^{(m)})$ where $\hat H(Y)$ is the empirical entropy of the verb prior over the full verb-labeled corpus. Conditional MI is the per-fold difference of two cross-entropies; its $5$-fold mean and standard error are reported in the main text. Wilcoxon signed-rank tests are computed on the $5$ paired samples.

\section{\texorpdfstring{$R^2$}{R2} Commonality Decomposition}
\label{app:r2_commonality}

As a corroborating analysis, we apply $R^2$ commonality (variance partitioning) on top of the same probe representations used in Section~\ref{sec:motion_goal}. We extract frozen $[\mathrm{CLS}]$ embeddings from the motion-only and goal-only probes, then regress one-hot verb labels on three feature sets --- motion only, goal only, and their concatenation --- using ridge regression. The decomposition separates variance uniquely explained by action trajectories, variance uniquely explained by endpoint state changes, and variance shared by both:
\begin{align*}
    R^2_{\mathrm{unique\;motion}} &= R^2_{\mathrm{cat}} - R^2_{\mathrm{vision}}, \\
    R^2_{\mathrm{unique\;goal}}   &= R^2_{\mathrm{cat}} - R^2_{\mathrm{action}}, \\
    R^2_{\mathrm{shared}}         &= R^2_{\mathrm{action}} + R^2_{\mathrm{vision}} - R^2_{\mathrm{cat}}.
\end{align*}
The concatenated representation explains $R^2_{\mathrm{cat}} = 0.551$ of verb variance. Of this, $R^2_{\mathrm{unique\;motion}} = 0.046$ is unique to action trajectories, $R^2_{\mathrm{unique\;goal}} = 0.122$ is unique to endpoint state, and the remaining $69.5\%$ of explained variance is shared. The qualitative pattern matches the mutual-information decomposition in Section~\ref{sec:motion_goal}: a small but reliable unique-motion component sits on top of a substantial shared component, with goal carrying somewhat more unique information than motion (here, ratio ${\approx}2.6\times$; in the MI estimate, ${\approx}4.8\times$).

\section{Per-Verb MI Contribution}
\label{app:per_class}

Table~\ref{tab:per_class_mi} reports the per-verb contribution to total mutual information for each modality condition, pooled over the five held-out folds. The contribution of class $c$ to total $I(Y; X^{(m)})$ is
\[
    \mathrm{contrib}_c(m) \;=\; \pi_c \,\bigl(-\log_2 \pi_c \;-\; \overline{\mathrm{CE}}_c(m)\bigr),
\]
where $\pi_c$ is the class prior and $\overline{\mathrm{CE}}_c(m)$ is the average cross-entropy on val episodes whose true verb is $c$. Per-modality $\Delta_{\mathrm{motion}} = \mathrm{contrib}_c(\text{Both}) - \mathrm{contrib}_c(\text{Goal})$ and $\Delta_{\mathrm{goal}} = \mathrm{contrib}_c(\text{Both}) - \mathrm{contrib}_c(\text{Motion})$ isolate where each modality's unique contribution lives; their column totals match the global conditional MIs reported in Section~\ref{sec:motion_goal} up to fold-pooling rounding.

\begin{table}[htbp]
\centering
\caption{Per-verb MI contribution (bits, pooled across $5$ CV folds). $\pi \log_2 \pi$ is the entropy contribution from the class prior.}
\label{tab:per_class_mi}
\small
\resizebox{\columnwidth}{!}{%
\begin{tabular}{lrrrrrrr}
\toprule
Verb & count & $\pi \log_2 \pi$ & Motion & Goal & Both & $\Delta_{\rm motion}$ & $\Delta_{\rm goal}$ \\
\midrule
move    & 27{,}291 & 0.528 & 0.157 & 0.189 & 0.213 & $+0.023$ & $+0.056$ \\
put     & 26{,}049 & 0.526 & 0.134 & 0.143 & 0.161 & $+0.018$ & $+0.027$ \\
fold    &  5{,}160 & 0.252 & 0.143 & 0.189 & 0.187 & $-0.001$ & $+0.044$ \\
place   &  4{,}671 & 0.236 & 0.056 & 0.072 & 0.079 & $+0.007$ & $+0.023$ \\
take    &  4{,}428 & 0.228 & 0.121 & 0.162 & 0.162 & $+0.000$ & $+0.041$ \\
unfold  &  3{,}903 & 0.209 & 0.100 & 0.145 & 0.156 & $+0.011$ & $+0.056$ \\
sweep   &  2{,}889 & 0.170 & 0.162 & 0.166 & 0.165 & $-0.001$ & $+0.003$ \\
open    &  1{,}896 & 0.126 & 0.097 & 0.105 & 0.099 & $-0.006$ & $+0.002$ \\
close   &  1{,}695 & 0.116 & 0.079 & 0.099 & 0.097 & $-0.001$ & $+0.018$ \\
pick up &  1{,}542 & 0.108 & 0.087 & 0.089 & 0.089 & $+0.000$ & $+0.002$ \\
turn    &     576 & 0.050 & 0.044 & 0.045 & 0.046 & $+0.001$ & $+0.002$ \\
remove  &     489 & 0.044 & 0.009 & 0.012 & 0.012 & $+0.001$ & $+0.003$ \\
reach   &     366 & 0.035 & 0.034 & 0.034 & 0.035 & $+0.000$ & $+0.000$ \\
push    &     357 & 0.034 & 0.011 & 0.008 & 0.014 & $+0.005$ & $+0.002$ \\
slide   &     240 & 0.025 & 0.008 & 0.007 & 0.010 & $+0.003$ & $+0.002$ \\
unzip   &     132 & 0.015 & 0.014 & 0.014 & 0.014 & $-0.000$ & $+0.000$ \\
cover   &     129 & 0.015 & 0.004 & 0.004 & 0.004 & $+0.000$ & $+0.000$ \\
\midrule
Total   & 81{,}813 & 2.717 & 1.260 & 1.483 & 1.542 & $+0.059$ & $+0.282$ \\
\bottomrule
\end{tabular}
}
\end{table}

\section{Artifact Licenses and Intended Use}
\label{app:licenses}

All datasets and pretrained models used in this work are publicly
released for academic research, and we use them in a manner consistent
with their stated intended use.

\paragraph{Datasets.}
BridgeV2~\citep{walke2023bridgedata} is released under the
Creative Commons Attribution 4.0 (CC-BY-4.0) license.
The dataset contains no personally identifying information or offensive
content: it consists of third-person camera frames of a robot arm
manipulating household objects, together with short
manipulation-instruction strings (e.g.\ ``put the pot on the stove'').

\paragraph{Pretrained models.}
The miniVLA base VLM checkpoint~\citep{belkhale2024minivla} and the
Stanford VQ-VAE bridge tokenizer are released under the MIT license by
the Stanford ILIAD group via HuggingFace.
Qwen2.5-0.5B~\citep{qwen2025qwen25} is released under the Apache 2.0
license.
DINOv2~\citep{oquab2024dinov2} is released under the Apache 2.0 license.
The FAST action tokenizer~\citep{pertsch2025fast} is released under the
Apache 2.0 license.
SimplerEnv~\citep{li2024simpler} is released under the MIT license.

\paragraph{Software.}
SpaCy is released under the MIT license. PyTorch is released under
a BSD-style license.

\paragraph{Derived artifacts.}
Our SALT tokenizer checkpoints and probe code are intended for
academic research only and will be released under a research-only
license.

\section{Compute Budget}
\label{app:compute}

\paragraph{Hardware.}
All training runs were performed on NVIDIA L40S GPUs (48~GB) on a
shared academic cluster.

\paragraph{Policy training.}
Each headline policy (SALT, VQ-VAE, and FAST miniVLA) was trained
for $15$k gradient steps at global batch size $128$ on $2$ L40S GPUs
for roughly $1$--$2$ days of wall-clock time per run.

\paragraph{Action tokenizer training.}
SALT tokenizers ($14$ configurations across the $n_g \times
n_{\rm emb}$ sweep) and the ablation tokenizers were each
trained on $1$ L40S GPU for $4$--$8$ hours (depending on codebook size),
for a total of roughly $150$ GPU-hours across all tokenizer runs.

\paragraph{Probes.}
Each $5$-fold verb-probe (Transformer-based) takes roughly $20$ minutes
on a single L40S; the full set of probes reported in Tables~1, 2, 4 and
Figure~3 totals ${\approx}80$ GPU-hours.

\paragraph{Total.}
Aggregate compute, including failed runs and ablation runs not reported
in the main text, is approximately $1{,}500$ L40S GPU-hours.

\section{Use of AI Assistants}
\label{app:ai_assistants}

AI coding assistants were used to
accelerate routine engineering tasks: writing plotting scripts, SLURM submit-script scaffolding,
LaTeX formatting, and proofreading the manuscript. All scientific contributions ---
research questions, experimental designs, claims,
analyses, and result interpretations --- are the authors'.

\end{document}